\documentclass[journal]{IEEEtai}

\usepackage[colorlinks,urlcolor=blue,linkcolor=blue,citecolor=blue]{hyperref}

\usepackage{color,array}

\usepackage{graphicx}

\usepackage{hyperref}
\newtheorem{Definition}{Definition}

\usepackage{booktabs}
\usepackage{multirow}
\usepackage{caption}
\usepackage{makecell} 
\usepackage{float}
\usepackage{color}
\usepackage{amsmath}
\usepackage{amssymb} 
\usepackage{mathrsfs}
\usepackage[ruled]{algorithm2e}
\usepackage{hyperref}
\usepackage{bm}
\newcommand{\vect}[1]{\bm{#1}}

\usepackage{graphicx}

\begin{document}

\title{DualHNIE: Dual-Channel Hypergraph Learning for Node Importance Estimation in Heterogeneous Knowledge Graphs}

\author{Jiawen Chen, Yanyan He, Qi Shao, Mengli Wei, Duxin Chen,~\IEEEmembership{Member, IEEE}, Wenwu Yu,~\IEEEmembership{Senior Member, IEEE}, and Yanlong Zhao,~\IEEEmembership{Senior Member, IEEE}
\thanks{This research was supported by the National Key Research and Development Program of China (Grant No.G2025YFF0524100), the National Natural Science Foundation of China (Grants No.62233004, 62273090, and T2541017), the Jiangsu Provincial Scientific Research Center of Applied Mathematics (Grant No.BK20233002), and the Basic Research Program of Jiangsu (Grants No.BK20253018 and BK20253020). (Corresponding authors: Duxin Chen, Wenwu Yu.) }
\thanks{Jiawen Chen, Qi Shao, Mengli Wei, Duxin Chen, and Wenwu Yu are with the Jiangsu Key Laboratory of Networked Collective Intelligence, School of Mathematics, Southeast University, Nanjing 210096, China. (e-mail:chenjiawen@seu.edu.cn; shaoqi@seu.edu.cn; weimengli@seu.edu.cn; chendx@seu.edu.cn; wwyu@seu.edu.cn).}
\thanks{Yanyan He is with the Jiangsu Key Laboratory of Networked Collective Intelligence, School of Cyber Science and Engineering, Southeast University, Nanjing 210096, China. (e-mail: heyy@seu.edu.cn).}
\thanks{Yanlong Zhao is with the State Key Laboratory of Mathematical Sciences, Academy of Mathematics and Systems Science, University of Chinese Academy of Sciences, Beijing 100190, China, and also with the School of Mathematical Sciences, University of Chinese Academy of Sciences, Beijing 100049, China (e-mail: ylzhao@amss.ac.cn) }
}

\markboth{IEEE Transactions on Artificial Intelligence, DOI: 10.1109/TAI.2026.3712336}
{Chen J. \MakeLowercase{\textit{et al.}}: DualHNIE}

\maketitle

\begin{abstract}
Estimating node importance in heterogeneous knowledge graphs is a fundamental problem underlying recommendation, search, and knowledge decision systems. However, most existing methods rely on pairwise message passing mechanisms that fail to capture higher-order interactions induced by meta-relational structures. Furthermore, structural topology and semantic attributes are typically entangled within a unified embedding space, which obscures their distinct inductive biases and limits the discriminative capacity of learned importance representations.
To address these limitations, we propose DualHNIE, a principled dual-channel hypergraph learning framework for node importance estimation. DualHNIE first constructs a higher-order knowledge graph by forming typed hyperedges from meta-path sequences, enabling explicit modeling of higher-order relational patterns. It then introduces two complementary encoders: a structure-aware hypergraph attention network that performs locally normalized aggregation over meta-path--induced hyperedges to capture localized structural dependencies, and a sparse--chunked hypergraph transformer that captures global semantic interactions while maintaining scalable computation. We further design a contrastive alignment mechanism with auxiliary supervision, ensuring cross-view consistency while preserving modality-specific representation. Extensive experiments on multiple benchmark datasets demonstrate that DualHNIE outperforms state-of-the-art methods, validating the effectiveness of explicit high-order modeling and disentangled dual-channel representation learning for heterogeneous knowledge graphs. Code and datasets are available\footnote[1]{\url{ https://github.com/jiawenchen10/DualHNIE}}.
\end{abstract}

\begin{IEEEImpStatement}
Accurate estimation of node importance in heterogeneous knowledge graphs is critical for recommendation systems, including personalized ranking, query understanding, and knowledge-aware reasoning.
Existing approaches struggle to capture high-order multi-entity interactions and often conflate structural and semantic signals, limiting their effectiveness on large, real-world graphs. DualHNIE introduces a dual-channel hypergraph architecture that models higher-order relations while disentangling structure and semantics through a sparse--chunked transformer and contrastive fusion. Experiments on four benchmarks achieve superior performance over baseline methods, and the proposed sparse--chunked attention module reduces runtime by over 30\% and memory usage by nearly 50\% on large scale heterogeneous
knowledge graphs, enabling more efficient deployment in recommendation systems and ranking retrieval. 
\end{IEEEImpStatement}


\section{Introduction}
\IEEEPARstart{K}nowledge Graphs (KGs) have emerged as a fundamental paradigm for representing complex, multi-typed real-world data~\cite{liang2024survey,li2025mask}. In contrast to homogeneous graphs with a single type of node and edge, heterogeneous KGs incorporate diverse entities, such as users, items, authors, and movies connected by rich semantic relations \cite{zhang2023autoalign}. 
Within this context, a critical problem is node importance estimation (NIE), which assigns importance scores to entities based on their structural and semantic roles in the recommendation and search systems \cite{park2019estimating,Chen2025CriticalNodes}, for example, Pagerank~\cite{page1999pagerank} assigns importance scores to visited pages on the World Wide Web. Accurate and efficient NIE is essential for numerous downstream applications~\cite{zhang2024knowgpt}, including recommendation systems and information retrieval.
Heterogeneous information in knowledge graphs serves as a crucial and richly diverse resource that enhances ranking and retrieval by capturing structure and entity semantics.

In heterogeneous knowledge graphs, the importance of an entity is determined not only by pairwise relations but also by collective behavioral and semantic patterns that emerge from interactions among multiple entities \cite{hayat2024heterogeneous}. Existing HKG models mostly rely on neighbor information, which capture only pairwise dependencies~\cite{shao2024unique,li2025multi}. 
However, most graph neural networks (GNNs) remain limited to pairwise message passing, which constrains their ability to capture true high-order correlations, such as users are connected with an item~\cite{yin2024high,xie2025contrastive}. 
Although hypergraphs offer a natural formalism for modeling such multi-entity interactions~\cite{zhang2025learning,leestructure}, current high-order approaches often resort to k-hop neighborhoods, decomposing complex group relations into sequences of pairwise interactions. This limitation restricts the model’s ability to jointly leverage heterogeneous semantics and intrinsic high-order structures, posing a key challenge for advancing knowledge-aware representation learning in HKGs.

{Existing approaches for heterogeneous knowledge graphs can be roughly grouped into structure-oriented and semantic-oriented modeling paradigms \cite{li2025multi,leestructure}. Structure-driven methods primarily estimate node importance based on topological signals such as in-degree~\cite{park2019estimating,park2020multiimport} and similarity~\cite{wang2025semsi} to estimate node importance, these methods concentrate solely on the semantic attributes of entities~\cite{chen2024deep,wang2019heterogeneous,10974738}. However, node importance is inherently governed by the interplay between structural position and semantic context. Although some recent works have attempted to integrate structural and semantic signals using attention mechanisms \cite{liu2025multi,chung2023representation}, these approaches usually employ dense matrix multiplications, leading to high computational complexity on the large scale knowledge graphs. Moreover, most current models tightly couple structural and semantic information~\cite{liu2025teaching}, which fails to explicitly disentangle and independently evaluate their complementary contributions to representation learning.}

{To address the above limitations, our proposed framework DualHNIE for node importance estimation in HKGs guided by three design principles: higher-order modeling, dual-view disentanglement, and efficient cross-view alignment.}
First, to capture higher-order dependencies, we construct a meta-path--induced hypergraph, where typed hyperedges connect all entities shared with the same relation, enabling explicit modeling of multi-entity interactions beyond pairwise HKGs.
Second, DualHNIE adopts a dual-channel architecture to disentangle structural and semantic features. A structure-aware hypergraph attention network aggregates local higher-order topology, while a semantic hypergraph transformer encodes global contextual information using a sparse chunking strategy to reduce the cost of dense attention. Finally, we fuse structural and semantic embeddings through contrastive alignment with auxiliary supervision, yielding robust representations that jointly exploit both information sources.
Our main contributions are summarized as follows: 
\begin{itemize}
 \item We propose a meta-path--induced heterogeneous higher-order hypergraph construction framework that explicitly models multi-entity and higher-order interactions, addressing the limitation of pairwise-only methods.
 \item We introduce a dual-channel encoding architecture that disentangles structural and semantic information. A structure-aware hypergraph attention network models higher-order topological relations, while a contextual hypergraph transformer encodes semantic dependencies, mitigating the expressiveness limitations of prior mixed or single-channel approaches.
 \item We develop a contrastive alignment and fusion mechanism that jointly optimizes structural and semantic embeddings under auxiliary supervision, enabling effective integration of heterogeneous information for accurate node importance estimation.

\end{itemize}

\section{Related Work}
In this section, we summarize previous research in the following two areas: node importance estimation and higher-order relational modeling in heterogeneous knowledge graphs.

Estimating node importance is pivotal in knowledge graph representation learning, yet existing methods face two key limitations: (i) they fail to capture high-order multi-entity interactions beyond pairwise relations, and (ii) they lack explicit disentanglement between structural topology and semantic context.
Single-view approaches generally employ GNNs with attention mechanisms. For instance, GENI~\cite{park2019estimating} employs predicate-aware attention with centrality adjustment, MultiImport~\cite{park2020multiimport} integrates edge-aware attention and clustering, and HIVEN~\cite{huang2022estimating} models heterogeneous relational structures. While effective in capturing local neighborhood signals~\cite{huang2025sparse}, these methods are inherently constrained by pairwise message passing.
Recent multi-view frameworks attempt to enrich representations~\cite{ma2025node} by fusing heterogeneous signals. LICAP~\cite{zhang2024label} leverages label-guided pre-training, CADReN~\cite{zhong2025cadren} introduces contextual anchors for cross-graph generalization, and EASING~\cite{chen2025semi} jointly models importance and uncertainty. However, these approaches typically couple structural and semantic information~\cite{liu2025teaching}, without explicitly separating their distinct roles. Critically, existing approaches neither model native high-order structures in HKGs nor offer a mechanism to disentangle and jointly align structural and semantic views.

Interactions in HKGs often involve multi-item or multi-entity dependencies that cannot be captured by pairwise graph structures. 
Standard GNNs encode only k-hop neighborhoods~\cite{yin2024high}, leaving complex semantic associations underrepresented. {To address this, N-ary relational modeling~\cite{fatemi2021knowledge} is introduced for richer group-level semantics. Self-supervised techniques have also been applied to generate higher-order relations using user labels~\cite{Khan3613964, xia2022self}, but these relations focus on higher-order relations in multi-hop neighborhoods, do not correspond to meta-path-derived higher-order dependencies. Hypergraph-based methods further involves interactions beyond pairwise connections as hyperedges. For example, ID-HAN incorporates hyperedge to enhance interaction modeling~\cite{chen2024hyperedge}.
Recent hypergraph neural network (HGNN)~\cite{gao2022hgnn+} studies explore dual-feature fusion~\cite{ju2024hypergraph, wang2024dual}, and some dual-channel HGNNs (e.g., DualHGNN~\cite{wang2021dualgnn}, DPHGNN~\cite{saxena2024dphgnn}, DVHGNN~\cite{li2025dvhgnn}) handle heterogeneous modalities. 
In most cases, they share similar HGNN backbones operators such that structural and semantic features are jointly encoded but not explicitly disentangled and not tailored to node importance estimation, which limits their ability to capture typed higher-order semantics in HKGs. Our framework constructs structurally and semantically channels based on meta-path--induced hyperedges and couples them with contrastive alignment specifically designed for node importance estimation. }

\section{{PRELIMINARIES and Background}}
In this section, we introduce the preliminaries of the node importance estimation and knowledge graphs.


\begin{Definition}[Heterogeneous Knowledge Graph]
A heterogeneous knowledge graph is denoted as \( \mathcal{G} = (\mathcal{V}, \mathcal{E}) \), where \( \mathcal{V} \) is a set of entities that may belong to multiple types (e.g., people, movies in Fig.~\ref{fig01}(a)(b)(c)), and \( \mathcal{E} \subseteq \mathcal{V} \times \mathcal{R} \times \mathcal{V} \) is a set of relational edges. Each edge is represented as a triplet \( (v_h, r_i, v_t) \), indicating that head entity \( v_h \) is connected to tail entity \( v_t \) via a relation \( r_i \in \mathcal{R} \), where \( \mathcal{R} \) denotes the set of relation types. The heterogeneity of \( \mathcal{G} \) arises from the coexistence of diverse entity types and relation types, enabling rich semantic representation of real-world knowledge.
\end{Definition}

\begin{Definition}[Meta-Paths]
{A meta-path $\mathcal{P}$ is a path pattern defined as a sequence of node entities and relation types in Fig.~\ref{fig01}(c), such as $v_1 \xrightarrow{r_1} v_2 \xrightarrow{r_2} \cdots \xrightarrow{r_{n}} v_{n+1}$.} 
\end{Definition} 

\begin{Definition}
\textit{(Heterogeneous Higher-order Knowledge Graphs)}:
we denote a heterogeneous higher-order knowledge graphs (HHKGs) as \(\mathcal{H} = (\mathcal{V}, \mathcal{\tilde{E}})\), which consists of nodes set \( \mathcal{V} \) and hyperedges set \(\mathcal{\tilde{E}}\). 
The structural features \( \mathcal{X}_1 \in \mathbb{R}^{|\mathcal{V}| \times d_1} \) and semantic features $\mathcal{X}_2 \in \mathbb{R}^{|\mathcal{V}|\times d_2}$, where \( d_1, d_2 \) are the dimension of feature space.
The hyperedge feature matrix is \( \mathbf{E} \in \mathbb{R}^{|\mathcal{E}| \times d'} \), where \( d' \) is the hyperedge feature dimension. 
\end{Definition} 

\begin{Definition}[Node Importance Estimation]
Given a heterogeneous higher-order knowledge graph \(\mathcal{H} = (\mathcal{V}, \mathcal{\tilde{E}})\), its structural features $\mathcal{X}_1$, semantic features $\mathcal{X}_2$ and hyperedges feature $E$, and a set of observed importance scores $s \in \mathbb{R}^{|\mathcal{V}_s|}$ assigned to a subset of nodes $\mathcal{V}_s \subseteq \mathcal{V}$,where the scores are derived from external signals such as \textit{popularity}, \textit{recommendation scores}. The task aims to learn a scoring function $f_\theta$ parameterized by $\theta$ that predicts importance $\hat{s}$ for all nodes:
\begin{equation}
 (\mathcal{V}, \mathcal{\tilde{E}}, \mathcal{X}_1, \mathcal{X}_2,\ \mathbf{E}) \xrightarrow{f_\theta(\cdot)} \hat{s} \in [0, \infty).
\end{equation}
\end{Definition}

\section{Methodology}
In this section, we provide a detailed elaboration of the proposed DualHNIE and its {components.}

\subsection{Overview}
{The proposed DualHNIE framework is illustrated in Fig.~\ref{fig03}, which introduces a dual-channel hypergraph representation learning to disentangle structure-semantics features with a structural--relational encoder and contextual hypergraph transformer, and then aligns them through a contrastive fusion module for node importance estimation.
Given an input HKG, we first construct the meta-path--induced hyperedges $\tilde{\mathcal{E}}$ from relational triples to model higher-order multi-entity interactions. In parallel, semantic prompts are constructed from dataset-provided descriptions and encoded as semantic node features $\mathcal{X}_2$ by a pretrained language model. Based on these two sources, we build a higher-order structural hypergraph and a semantic hypergraph, which serve as inputs to the two encoding channels shown in Fig.~\ref{fig03}(a)(b). Finally, we formulate a Heterogeneous Higher-order Knowledge Graphs 
into the high-order structural and semantic knowledge graphs into dual-channel input in Fig.~\ref{fig03}(c).
The structural channel employs a structure-aware hypergraph attention encoder to aggregate higher-order relations and produce structure-level embeddings. The semantic channel uses a sparse-chunked hypergraph Transformer to model contextual dependencies from prompt-derived semantic features with reduced attention cost. Finally, the two channel embeddings are mapped into a shared space and integrated through a cross-channel contrastive alignment and adaptive fusion module, which enforces cross-view alignment and produces the final representations used for importance estimation.
}

\begin{figure}[htbp]
 \centering
 \setlength{\abovecaptionskip}{0pt}
 \setlength{\belowcaptionskip}{0pt} 
 \includegraphics[width=1\linewidth]{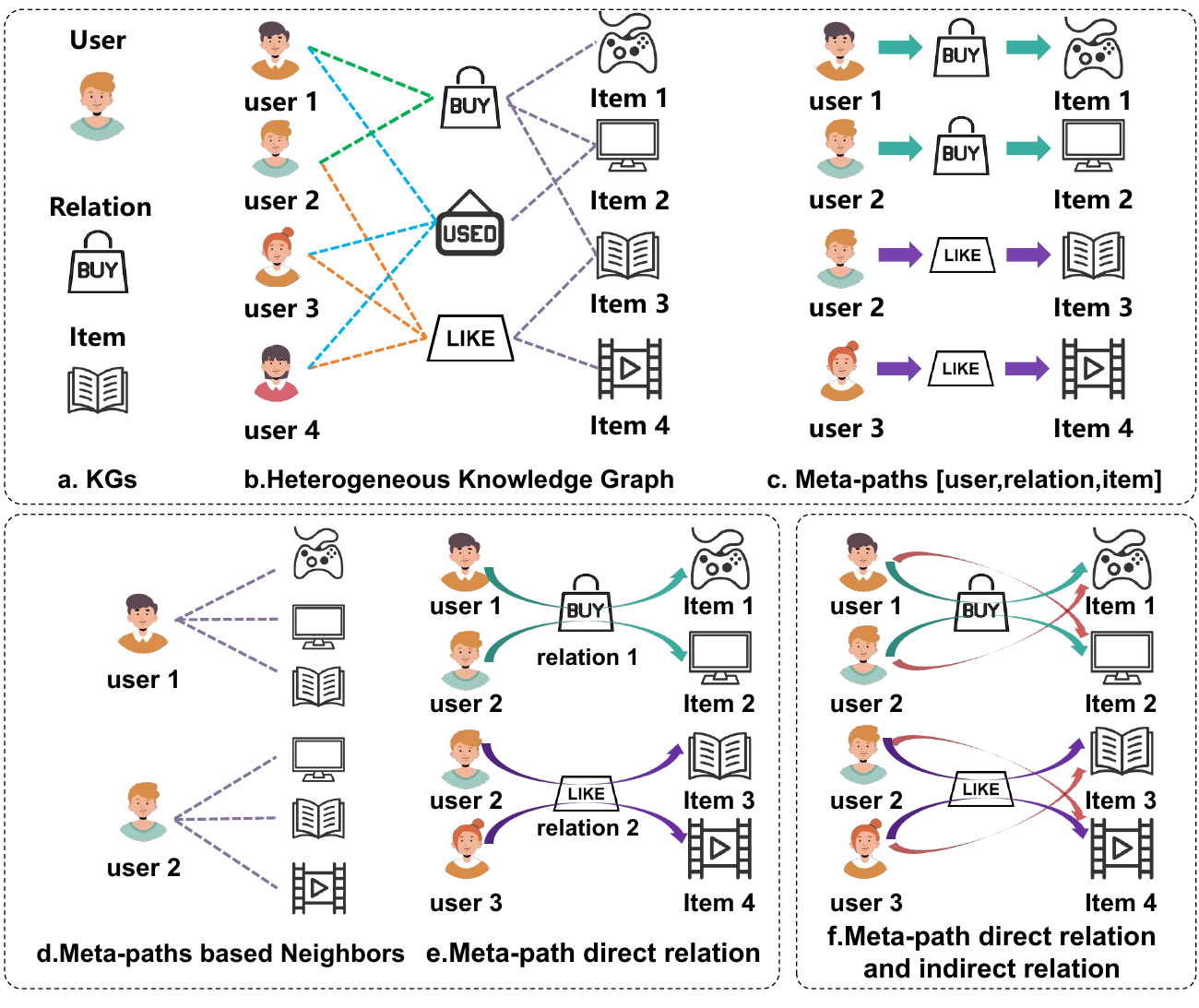}
 \caption{Illustration of HKGs.
 (a) Knowledge graph elements. (b) Heterogeneous knowledge graph. 
 (c) Meta-path examples (i.e., user-relation-item). (d) Users and their neighbors based on multiple meta-path. (e) Meta-path direct relations with its neighbors. (f) {Meta-path direct relation and indirect relations (red arrow), e.g., an indirect meta-path as [user$_1$, buy, item$_2$].}}
 \label{fig01}
\end{figure}

\subsection{Heterogeneous Higher-order Knowledge Graphs}
Existing modeling of higher-order roles in knowledge graphs relies on neighbors (Fig.~\ref{fig01}(d)(e)), which essentially emphasizes only direct connections~\cite{yin2024high}. However, indirect but unobserved and latent links (e.g., Fig.~\ref{fig01}(f)) are often overlooked, which reveal hidden associations between different users and items.
To address this, we introduce the Heterogeneous Higher-order Knowledge Graphs based on meta-path relation, which groups all users and items shared with the same relation type into a relational hyperedge to describe the collective associations induced by that relation.

\textbf{Higher-order Structural Knowledge Graph}.{ Given a collection of observed relational triples in the form of $(u, r, i)$, where $u \in \mathcal{U}$ represents a user, $i \in \mathcal{I}$ represents an item, and $r \in \mathcal{R}$ denotes the relation type, we define a hyperedge $e_r$ corresponding to multi-relation $r$ as $
e_r = \{u_{1}, u_{2}, \cdots, u_{n}, i_{1}, i_{2}, \cdots, i_{m}\}$. 
Each $(u_j, r, i_k)$ is an observed triple with relation $r$. The hyperedge $e_r \in \tilde{\mathcal{E}}$ aggregates all users and items that are associated for the hypergraph construction. 
Based on this, we introduce the notion of the higher-order meta-paths for relational hyperedge as} 
\begin{Definition}[Higher-order Meta-path]
{In HKGs, a higher-order meta-path associated with a relation type $r\in \mathcal{R}$ represents the set of entities connecting with relation $r$, It is defined as $\mathcal{\hat{P}}(r_{j_pk_q})=\{u_{j_{1,p}},\dots,u_{j_{k,p}},i_{k_{1,q}},\dots,,i_{k_{k,q}}\}$, which induces a hyperedge $e_r$ containing all users and items linked by $r$.} 
\end{Definition}

To preserve this heterogeneity of relation in HHKGs, we construct new hyperedge feature $E$ by concatenating original triples. Each hyperedge is thus assigned a derived type and embedded into a feature tensor, which incorporates heterogeneous relational semantics into the hypergraph and enriches its context. 
To characterize the derived type of hyperedges, all node pairs of a hyperedge $e_r$ formed as $\mathcal{P}(e_r) = \{(u_{j_p}, i_{k_q}) \mid u_{j_p} \in e_r, i_{k_q} \in e_r \}$. Here, triple $(u_j, r, i_k)$ is associated with an edge type $r_{jk} \in \mathcal{R}$. The set of edge types is then given by:
\begin{equation}
\label{eq02}
\mathcal{T}(e_r) = \{r_{j_p k_q} \mid (u_{j_p}, r_{j_p k_q}, i_{k_q}) \in \text{triples} \}.
\end{equation}
To derive the feature representation of each hyperedge, we remove redundant relation types from $\mathcal{T}(e_r)$ and encode the remaining ordered types as a tuple
$\mathbf{t}_{e_r} = [\, r_{j_1k_1},\, r_{j_2k_2},\, \ldots,\, r_{j_lk_l} \,]$ ,
where $l = |\mathcal{T}(e_r)|$. Each distinct tuple $\mathbf{t}_{e_r}$ is then assigned a unique identifier. 
By collecting the encoded tuples for all $e_r \in \mathcal{E}$, we construct the hyperedge feature tensor $\mathbf{E} \in \mathbb{R}^{|\mathcal{E}| \times d'}$. Therefore, we obtain a high-order structural knowledge graph $\mathcal{H}_1(\mathcal{V},\mathcal{\tilde{E}},\mathcal{X}_1,\textbf{E})$.

\begin{figure*}
 \centering
\includegraphics[width=0.95\linewidth]{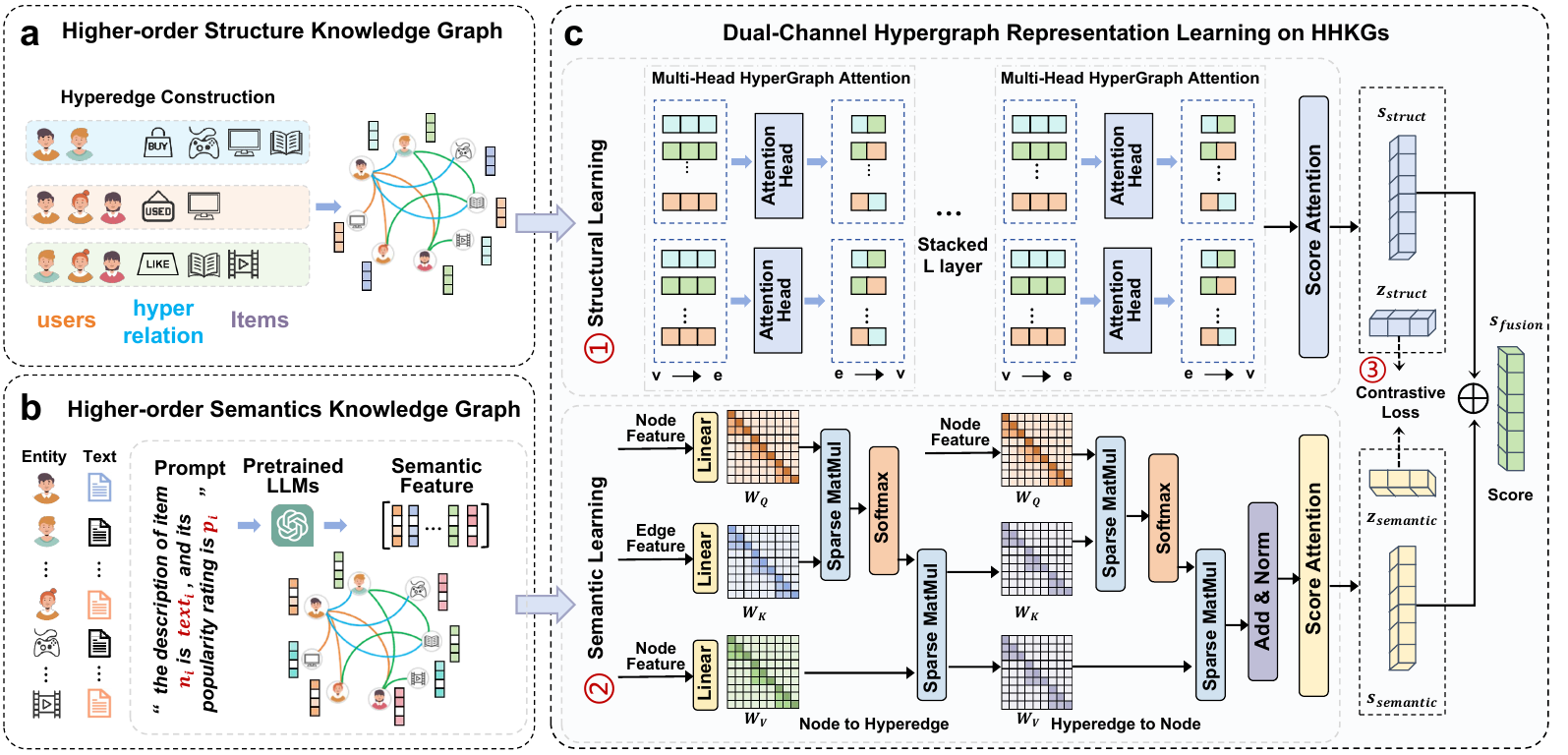}
\caption{Overview of the proposed DualHNIE framework.
{
(a) Higher-order structural hypergraph constructed from meta-path--induced group relations.}
(b) Higher-order semantic hypergraph representing contextual patterns with encode-only pretrained LLM(BERT).
(c) Dual-channel encoder: $\bigcirc\!\!\!\!1$ structure-aware hypergraph attention network,
$\bigcirc\!\!\!\!2$ semantic hypergraph transformer with sparse--chunked attention, and
$\bigcirc\!\!\!\!3$ contrastive alignment and fusion for final node importance estimation.}
 \label{fig03}
\end{figure*}

\textbf{Higher-order Semantic Knowledge Graph.}
{Given the semantic information $\mathscr{D} = \{d_i\}_{i=1}^N$, each node $v_i \in \mathcal{V}$ is associated with a textual name $n_i$, a textual description $d_i$, and an item popularity score $p_i \in \mathbb{R}$. The descriptions $d_i$ are sourced from the original datasets (e.g., entity descriptions on FB15K) and are not generated by large language model. To facilitate semantic encoding, we construct a structured prompt template for each node as 
``the description of item $n_i$ is $d_i$, and its popularity rating is $p_i$.''
The prompt textual input is then encoded with a pretrained language model $\mathcal{M}$ to obtain the semantic embeddings:}
\begin{equation}
\label{equation03}
\mathbf{e}_i^{\text{sem}} = \mathcal{M}\big( f_{\text{concat}}(n_i, d_i, p_i) \big) \in \mathbb{R}^d,
\end{equation}
{where $f_{\text{concat}}(\cdot)$ denotes string concatenation and $\mathcal{M}$ is BERT pretrained model~\cite{JacobDevlin04805}, here BERT is an encoder-only model that does not perform prompt generative tasks. Specifically, after feeding the concatenated prompt sequence into BERT, we extract the semantic vector by taking the hidden state corresponding to the special [CLS] token from the final encoder layer. To maintain computational efficiency and preserve the pretrained language knowledge, the weights of the BERT model $\mathcal{M}$ are frozen during the training of the DualHNIE framework. The final semantic features $\mathcal{X}_2$ are then obtained via a multilayer perceptron (MLP) layer, defined as $\mathcal{X}_2 = \text{MLP}\big( [\mathbf{e}_1^{\text{sem}}, \dots, \mathbf{e}_N^{\text{sem}}]^\top \big)$ in Fig.~\ref{fig03}~(b).
We formulate higher-order semantic knowledge graph as $\mathcal{H}_2(\mathcal{V},\mathcal{\tilde{E}},\mathcal{X}_2,\textbf{E})$. Algorithm \ref{alg:alg1} displays the detailed construction of the dual-channel Heterogeneous Higher-order Knowledge Graphs.}

\begin{algorithm}[htbp]
\setlength{\baselineskip}{0.8\baselineskip}
\small 
\caption{Construction of HHKGs}
\label{alg:alg1}
\KwIn{Heterogeneous knowledge graph \( \mathcal{G} = (\mathcal{V}, \mathcal{E}, \mathcal{X}_1, \mathcal{D}) \)}
\KwOut{Heterogeneous higher-order structural and semantic knowledge graph $\mathcal{H}_1$, $\mathcal{H}_2$}

\BlankLine
\For{each relation $r\in \mathcal{R}$}{
 Collect users $\{u_j\}$ and items $\{i_k\}$ linked by $r$\;
 
  $e_r\leftarrow$ Relational Hyperedge $\{u_1,\ldots,u_n, i_1,\ldots,i_m\}$\;
 
 $\mathcal{T}(e_r)\leftarrow$ Relation Types from Triples by Equation.~(\ref{eq02}) \;
 
 Encode $\mathcal{T}(e_r)$ as $\mathbf{t}_{e_r}$, 
 Add $e_r$ to $\mathcal{\tilde{E}}$ and $\mathbf{t}_{e_r}$ to $\mathbf{E}$\;
}
Collect the hyperedges $e_r$ and construct structural knowledge higher-order graph $\mathcal{H}_1(\mathcal{V},\tilde{\mathcal{E}},\mathcal{X}_1,\mathbf{E})$\;

\BlankLine

\For{each node $v\in\mathcal{V}$ and description $d_i\in \mathcal{D}$}{
$\mathbf{E}^{\mathrm{sem}}\leftarrow$ Semantic Embeddings by Equation.~(\ref{equation03})\;
}
$\mathcal{X}_{2}\leftarrow$ Semantic Features as $ \mathrm{MLP}(\mathbf{E}^{\mathrm{sem}})$\;

Construct semantic knowledge higher-order graph $\mathcal{H}_2(\mathcal{V},\tilde{\mathcal{E}},\mathcal{X}_2,\mathbf{E})$\;

\end{algorithm}



\subsection{Hypergraph Structural Representation Learning} 
To learn heterogeneous higher-order graph and structural features in HHKGs $\mathcal{H}(\mathcal{V},\mathcal{\tilde{E}},\mathcal{X}_1,\textbf{E})$, we introduce a Hypergraph Attention (HGAT) to model multi-entity interactions for node importance estimation. HGAT provides a two-stage information method, which propagates messages from nodes to relation hyperedges and then from hyperedges back to nodes.

The $l$-th layer contains $d^{(l)}$ hypergraph score aggregation (HSA) heads. Let $s_h^{(l-1)}(i)$ and $s_h^{(l-1)}(e)$ denote the scores of node $i$ and hyperedge $e$ from the $(l-1)$-th layer, which are fed into the $h$-th HSA head in the $l$-th layer. This head aggregates the scores to update node score $s_h^{(l)}(i)$. 
Each layer $l$ employs $d^{(l)}$ attention heads to estimate node structural scores. For head $h$, an initial score is computed by a learnable function $\text{HSA}_h(\cdot)$:
$s_h^{(0)}(i) = \text{HSA}_h(\mathbf{z}_i)$,
where $\mathbf{z}_i$ is the structural feature of node $v_i$ from $\mathcal{X}_1$. 
The initial representation is formed by concatenating all heads:
\begin{equation}
\label{equation08}
 s_{\text{struct}}^{(0)} = \Big\Vert_{h=1}^{d^{(0)}} s_h^{(0)}(i),
\end{equation}
where $\Vert$ denotes concatenation.
For a hyperedge $e$ with incident nodes $\mathcal{V}(e)$, the aggregated score is computed as:
\begin{equation}
\small \label{eq004}
s_h^{(l)}(e) = \sum_{i \in \mathcal{V}(e)} \alpha_{e i}^{h, l} \cdot s_h^{ (l-1)}(i),
\end{equation}
where $\alpha_{e i}^{h, l}$ is the attention weight between node $i$ and hyperedge $e$. Relation between the intermediate scores of node $i$ and hyperedge $e$, and the roles played by the intermediate predicate are captured by the hypergraph attention layer.
This weight is computed as follows:
\begin{equation}
\small
\label{eq005}
\begin{aligned}
\mathcal{A}_{e i}^{h, l} &= \sum_m \vect{a}_{h, l}^\top \left[ s_h^{(l-1)}(i) \parallel \phi(p_{e i}) \parallel s_h^{(l-1)}(e) \right], \\
\alpha_{e i}^{h, l} &= \frac{\exp\left(\sigma_a\left(\mathcal{A}_{e i}^{h, l}\right)\right)}{\sum\limits_{k \in e } \exp\left(\sigma_a\left(\mathcal{A}_{e k}^{h, l}\right)\right)},
\end{aligned}
\end{equation}
where $\parallel$ denotes concatenation, $\phi(\cdot)$ is a learnable transformation applied to the predicate $p_{e i}$ between node $i$ and hyperedge $e$, $\vect{a}_{h,l}$ is the learning parameter, and $\sigma_a$ denotes the LeakyReLU function. The node score updates from all the connected hyperedges.
For each node $i$, information from its incident hyperedges $\mathscr{E}(i)$ is aggregated as:
\begin{equation}
\small
\label{eq006}
s_h^{(l)}(i) = \sum_{e \in \mathscr{E}(i)} \alpha_{i e}^{h, l} \cdot s_h^{(l)}(e),
\end{equation}
with the attention coefficient $\alpha_{i e}^{h, l}$ computed by:
\begin{equation}
\small
\label{eq007}
\begin{aligned}
\mathcal{A}_{i e}^{h, l} &= \sum_m \vect{a}_{h, l}^\top \left[ s_h^{(l)}(e) \parallel \phi(p_{i e}) \parallel s_h^{(l-1)}(i) \right], \\
\alpha_{i e}^{h, l} &= \frac{\exp\left(\sigma_a\left(\mathcal{A}_{i e}^{h, l}\right)\right)}{\sum\limits_{f \in \mathscr{E}_i} \exp\left(\sigma_a\left(\mathcal{A}_{i f}^{h, l}\right)\right)}.
\end{aligned}
\end{equation}
Subsequently, hyper-relational information is encoded through $L$ layers of hypergraph score attention. HGAT layer further applies a residual connection, layer normalization LN$(\cdot)$, and a feedforward network FFN$(\cdot)$:
\begin{equation}
\small
\label{eq09}
\begin{aligned}
\mathbf{s}^{(l)}_{\text{struct}} & = \text{HGAT}^{(l)}(\mathcal{H}, \mathbf{s}^{(l-1)}_{\text{struct}}), \\ 
\mathbf{s}^{(l)}_{\text{struct}} & = 
\text{Norm}\!\left(\mathbf{s}^{(l)}_{\text{struct}} + \text{FFN}(\mathbf{s}^{(l-1)}_{\text{struct}})\right).
\end{aligned} 
\end{equation}

The average pooling layers are employed with multi score attention heads and obtain the final representations as
\begin{equation}
\small
\label{eq10}
\mathbf{s}^{(L)}_{\text{struct}}=
\text{AVERAGE}\!\left(\mathbf{s}^{(L)}_{\text{struct}}\right).
\end{equation}
Then the structural representation of embedding and prediction scores with a linear output layer as
\begin{equation}
\label{eq11}
\vect{z}_{\text{struct}} = \vect{s}^{(L)}_{\text{struct}}, \quad \text{s}_{\text{struct}} = \text{FFN}(\vect{s}^{(L)}_{\text{struct}}).
\end{equation}

\subsection{Scalable Hypergraph Semantic Representation Learning}
{In our framework, we introduce a Sparse-Chunked Hypergraph Transformer (SAHGT) comprising $L$ stacked layers to enable scalable contextual modeling in higher-order semantic knowledge graph $\mathcal{H}(\mathcal{V},\mathcal{\tilde{E}},\mathcal{X}_2,\textbf{E})$. 
SAHGT incorporates the sparse local attention mechanism innovation and block-wise coordinated optimization strategy. 
Standard self-attention \cite{chung2023representation} 
computes attention scores for all node--hyperedge pairs, $Q_v K_e^\top / \sqrt{d}$, where $Q_v$ and $K_e$ are the projected query and key vectors for node $v$ and hyperedge $e$, respectively. This results in a computational complexity of $O(NMd)$, which becomes prohibitive for large-scale HHKGs ($N, M \gg 10^4$, e.g., N and M are the numbers of nodes and hyperedges). }

 {To tackle high computational complexity,
the dense incidence matrix $\mathbf{H}$ is stored in coordinate list (CL) format: $\mathrm{CL}(\mathbf{H}) = {(v_i, e_i)}_{i=1}^{\mathrm{nnz}(\mathbf{H})}$, where $\mathrm{nnz}(\cdot)$ is the number of non-zero entries. Attention scores are computed only over non-zero pairs and normalized via a softmax over hyperedges, updating hyperedge messages by aggregating from incident nodes:}
\begin{equation}
 h_e = \sum_{v \in \mathcal{N}(e)} 
 \sigma_b \!\left( Q_{v_i} K_{e_i}^\top /\sqrt{d} \right) 
 \cdot V_{v_i}, 
\end{equation}
 { where $i=1,\dots,\mathrm{nnz}(H)$ and $\sigma_b$ denotes the scatter softmax, $V_{v_i}$ is the value vector of node feature. 
We process the sparse indices in chunks of size $C$, enabling training on hypergraphs with millions of nodes and hyperedges. This chunked strategy is purely an implementation-level optimization for hardware efficiency on large-scale knowledge graphs.}


For each hyperedge $e$, message aggregates from its incident nodes $v$. Let $\mathbf{W}_Q^{(l)}$, $\mathbf{W}_K^{(l)}$, $\mathbf{W}_V^{(l)}$ be projection matrices at layer $l$. SAHGT computes the sparse attention weight from $v$ to $e$ and update the hyperedge representation
\begin{equation}
\small 
\label{eq013}
\begin{aligned}
 \alpha_{v \rightarrow e}^{(l)} & = \sigma_b \left( ( W_Q^{(l)} h^{(l-1)}_v ) \cdot ( W_K^{(l)} h^{(l-1)}_e ) /\sqrt{d} \right), \\
 h_e^{(l)} & = \sum_{v \in e} \alpha_{v \rightarrow e}^{(l)} \cdot W_V^{(l)} h^{(l-1)}_v,
\end{aligned}
\end{equation} 
followed by multi-head concatenation and a linear output projection.
Symmetrically, each node $v$ yields the aggregated message from its incident hyperedges 
\begin{equation}
\small
\label{eq14}
\begin{aligned}
 \alpha_{e \rightarrow v}^{(l)} & = \sigma_b\left( ( W_Q^{(l)} h^{(l-1)}_v ) \cdot ( W_K^{(l)} h^{(l-1)}_e ) / \sqrt{d} \right), \\
 h_v^{(l)} & = \sum_{e \in \mathcal{E}_v} \alpha_{e \rightarrow v}^{(l)} \cdot W_V^{(l)} h^{(l)}_e.
\end{aligned}
\end{equation} 

Then update with residual connections, batch normalization(BN) and a feedforward network with GELU activation:
\begin{equation}
\small
\label{eq15}
 s_{\mathrm{semantic}}^{(l)} = \mathrm{BN}_2 \!\left( s_{\mathrm{semantic}}^{(l-1)} + \mathrm{FFN}\!\left( \mathrm{BN}_1\!\left(s_{\mathrm{semantic}}^{(l-1)} + h_v^{(l-1)}\right) \right) \right),
\end{equation} 

By stacking $L$ SAHGT layers, the model iteratively updates node representations as
\begin{equation}
\small
 s^{(l)}_{\mathrm{semantic}} = \mathrm{SAHGT}^{(l)}(\mathcal{H}, s^{(l-1)}_{\mathrm{semantic}}), 
\end{equation}
\small
where $s^{(0)}_{\mathrm{semantic}} =\mathcal{X}_2$ are semantic features. The semantic representation 
prediction scores are obtained as
\begin{equation} 
\label{eq17}
 z_{\mathrm{semantic}} = s^{(L)}_{\mathrm{semantic}},\quad 
 s_{\mathrm{semantic}} = \mathrm{FFN}(s^{(L)}_{\mathrm{semantic}}).
\end{equation}

\subsection{Fusion Mechanism and Training}
To integrate structural and semantic representations, we propose a gated adaptive fusion module that dynamically balances the contributions of the two modalities. Specifically, the final node importance score $\hat{s}$ is computed as
\begin{equation}
\label{fusion}
 \hat{s} = \eta_{1} \odot s_{\mathrm{struct}} + \eta_{2} \odot s_{\mathrm{semantic}},
\end{equation}
where $\odot$ denotes element-wise multiplication, and
$\eta_{1}, \eta_{2} \ge 0,\eta_{1}+\eta_{2}=1$ is a learnable gating weight that adaptively modulates the influence of structural and semantic signals.

We apply regression supervision to three prediction branches: the structure-based output, the semantic-based output, and the fused output. Let $\mathcal{V}' \subseteq \mathcal{V}$ be the set of training (or validation) node indices, and $s$ denote the ground-truth target value. The Mean Squared Error (MSE) loss of fusion predictive score is defined as:
\begin{equation}
 \mathcal{L}_{\mathrm{fusion}} = \| \hat{s} - s \|_2.
\end{equation}

To enhance consistency between the structural and semantic feature spaces, $\mathbf{z}{\mathrm{struct}}, \mathbf{z}{\mathrm{semantic}} \in \mathbb{R}^{N \times H}$, we further incorporate a cross-modal contrastive learning objective. The cross-modal similarity matrix is computed as $\mathcal{S} = z_{\mathrm{struct}} \cdot z_{\mathrm{semantic}}^\top/\tau$, where $\tau>0$ is a temperature parameter. Here, positive pairs correspond to the embeddings of the same entity across $z_{\mathrm{struct}}$ and $z_{\mathrm{semantic}}$, while negative pairs are formed by embeddings of different entities.
The contrastive loss is formulated as the symmetric average of two cross-entropy terms: 
\begin{equation}
\small
 \mathcal{L}_{\mathrm{1}} 
 = -\frac{1}{2N} \sum_{i=1}^{N} \left[ 
\log \frac{\exp(\mathcal{S}_{i,i})}{\sum_{j=1}^{N} \exp(\mathcal{S}_{i,j})}
+ 
\log \frac{\exp(\mathcal{S}_{i,i})}{\sum_{j=1}^{N} \exp(\mathcal{S}_{j,i})}
\right] ,
\end{equation}
which encourages alignment of the two representation spaces.

To preserve the predictive capacity of each modality, we additionally supervise each prediction with the auxiliary loss
\begin{equation}
 \mathcal{L}_{2} = \frac{1}{2} \big [ \| s_{\text{struct}} -s \| ^2_2 + \| s_{\text{semantic}} -s \|^2_2 \big ].
\end{equation}

Finally, the overall loss function of DualHNIE is established as 
\begin{equation}
\label{fusionloss}
 \mathcal{L} = \mathcal{L}_{\mathrm{fusion}} 
 + \alpha \cdot \mathcal{L}_1 + \beta \cdot \mathcal{L}_2 ,
\end{equation}
where $\alpha, \beta \geq 0$ are hyperparameters that control the contributions of the contrastive and unimodal losses. The detailed process of the DualHNIE is shown in Algorithm.~\ref{algorithm01}.

\subsection{Complexity Comparison}\label{appendixb}
We analyze the computational complexity of DualHNIE and compare it against representative hypergraph attention models.
Classical hypergraph attention layers require computing node--hyperedge interactions for all incident pairs, leading to a complexity of $O(NEHd)$, where $N$ is the number of nodes, $E$ the number of hyperedges, $H$ the number of attention heads, and $d$ the head dimension.
Transformer-style hypergraph models typically construct a dense incidence matrix and evaluate attention over all node--hyperedge pairs. This incurs a computational cost of $O(NEd)$ and memory consumption of $O(NE)$, which becomes prohibitive for large hypergraphs. 
To improve scalability, SAHGT performs attention on sparse neighborhoods and avoids explicit dense incidence construction, reducing the complexity to $O(\mathrm{nnz}(\mathbf{H})Hd)$, where $\mathrm{nnz}(\mathbf{H})\ll NE$ for most real-world datasets. However, SAHGT still processes the entire sparse coordinate tensor as a monolithic block, which requires large contiguous index buffers and may exceed the CUDA $2^{31}-1$ indexing limit on large-scale hypergraphs.
DualHNIE addresses these limitations using a Sparse--Chunk aggregate module that preserves sparse computation while enabling chunk-wise coordinate processing. Attention is computed only on nonzero entries, keeping the overall complexity at $O(\mathrm{nnz}(\mathbf{H})Hd)$. In addition, restricting the softmax to per-hyperedge local neighborhoods reduces its cost from $O(NE)$ to $O(\mathrm{nnz}(\mathbf{H})\log k)$, where $k$ denotes the average hyperedge size. This chunked sparse strategy ensures both computational efficiency and hardware compatibility when handling large hypergraphs.

\begin{algorithm}[htbp]
\setlength{\baselineskip}{0.9\baselineskip}
\small 
\caption{The DualHNIE Framework}
\label{algorithm01}
\KwIn{Heterogeneous higher-order structural and 
semantic knowledge graph $\mathcal{H}_1$,$\mathcal{H}_2$}
\KwOut{$\hat{s} \leftarrow$ Final importance estimation score}
\BlankLine
\For{each epoch}{
  $s_{\mathrm{struct}}^{(0)}\leftarrow$ HGAT Initialization with $\mathcal{X}_1$ by Equation.~(\ref{equation08})\; 

 \For{$l=1$ \KwTo $L$ \textbf{and} head $h=1$ \KwTo $H$}{
  \If{$l < L$}{
   $s^{(l)}_h(e)\leftarrow$ Compute Hyperedges Embedding by Equation.~(\ref{eq004}),(\ref{eq005})\;
   
   $s^{(l)}_h(i)\leftarrow$ Compute Node Embedding by Equation.~(\ref{eq006}),(\ref{eq007})\;
  }
  \Else{
   $\mathbf{s}^{(l)}_{\text{struct}}\leftarrow$ Structural Importance score by Equation.~(\ref{eq09}),(\ref{eq10})\;
  }
 }
 Output $\mathbf{z}_{\text{struct}}$ and $\mathbf{s}_{\text{struct}}$ by Equation.~(\ref{eq11})\;

 \BlankLine

 $s_{\mathrm{semantic}}^{(0)} \leftarrow$ 
 SAHGT Initialization with $\mathcal{X}_2$\;
 

 \For{$l = 1$ \KwTo $L$}{
  \For{each chunk $c$}{
  \If{$l < L$}{
   $h_e^{(l)}\leftarrow$ Calculate Sparse-chunked Hyperedge Embedding by Equation.~(\ref{eq013})\;
   
   $h_v^{(l)}\leftarrow$ Calculate Sparse-chunked Node Embedding by Equation.~(\ref{eq14})\;
  }
  \Else{ $s^{(l)}_{\mathrm{semantic}}\leftarrow$ Semantic Importance score by Equation.~(\ref{eq15})\;
  }
   
  }
 }

 Output $\mathbf{z}_{\mathrm{semantic}}$ and $\mathbf{s}_{\mathrm{semantic}}$ by Equation.~(\ref{eq17})\;

 \BlankLine

 $\hat{s}\leftarrow$ Final Node Importance Estimation by Equation.~(\ref{fusion}). \;
 
 Update all learnable parameters by minimizing loss in Equation.~(\ref{fusionloss}) by Adam optimizer. 
}

\end{algorithm}

\section{Experiments}
In this section, we aim to validate the proposed framework DualHNIE through extensive experimental studies.

\begin{table}[h]
\centering
\setlength{\tabcolsep}{4pt} 
\caption{Statistics Details of HKGs and HHKGs}
\label{datasets}
\begin{tabular}{llcccc}
\toprule
\toprule
Type & Metric & FB15K & TMDB5K & IMDB & MUSIC10K \\
\midrule
\multirow{3}{*}{HKGs} 
  & Nodes  & 14,951 & 114,805 & 1,567,045 & 22,985 \\
  & Edges  & 592,213 & 761,648 & 14,067,776 & 65,290 \\
  & Relations & 1,345 & 34  & 28  & 8 \\
\midrule
\multirow{3}{*}{HHKGs} 
  & Hyperedges & 11,054 & 563  & 2,548  & 7 \\
  & Avg. Size & 1.3525 & 203.9165 & 441.5208 & 3,283.5714 \\
  & Max. Size & 240  & 51,958 & 190,095 & 9,999 \\
\bottomrule
\bottomrule
\end{tabular}

\end{table}

\begin{table*}[h]
\small 
\centering
\setlength{\tabcolsep}{4pt} 
\caption{Performance Comparison on Datasets in terms of Spearman correlation and NDCG@k (mean $\pm$ std). {Methods marked with the label $^{\dagger}$ utilize both structural and identical semantic information of DualHNIE.}}
\label{table 02}
\begin{tabular}{lccccccccc}
\toprule\toprule
 & \multicolumn{2}{c}{FB15K} & \multicolumn{2}{c}{TMDB5K} & \multicolumn{2}{c}{IMDB} & \multicolumn{2}{c}{MUSIC10K} \\
\cmidrule(lr){2-9}
 & SPEARMAN & NDCG@100 & SPEARMAN & NDCG@100 & SPEARMAN & NDCG@100 & SPEARMAN & NDCG@100 \\
\midrule
PPR & 0.350$\pm$0.019 & 0.841$\pm$0.011 & 0.686$\pm$0.010 & 0.850$\pm$0.008 & 0.648$\pm$0.006 & 0.876$\pm$0.020 & 0.189$\pm$0.023 & 0.798$\pm$0.011 \\
HAR & 0.202$\pm$0.012 & 0.826$\pm$0.005 & 0.630$\pm$0.009 & 0.814$\pm$0.021 & 0.632$\pm$0.005 & 0.795$\pm$0.036 & 0.177$\pm$0.019 & 0.799$\pm$0.014 \\
\hline
GCN & 0.466$\pm$0.029 & 0.878$\pm$0.015 & 0.659$\pm$0.047 & 0.862$\pm$0.023 & 0.720$\pm$0.015 & 0.895$\pm$0.015 & 0.444$\pm$0.033 & 0.885$\pm$0.020 \\
GraphSAGE & 0.753$\pm$0.009 & 0.929$\pm$0.008 & 0.547$\pm$0.032 & 0.830$\pm$0.032 & 0.705$\pm$0.020 & 0.880$\pm$0.015 & 0.459$\pm$0.020 & 0.859$\pm$0.011 \\
GENI & 0.771$\pm$0.009 & 0.917$\pm$0.014 & 0.735$\pm$0.022 & 0.852$\pm$0.013 & 0.755$\pm$0.010 & 0.915$\pm$0.010 & 0.493$\pm$0.026 & 0.887$\pm$0.007 \\
LICAP & 0.768$\pm$0.013 &  \underline{0.941$\pm$0.009} & 0.731$\pm$0.018 & 0.885$\pm$0.020 & 0.760$\pm$0.012 & 0.920$\pm$0.010 & 0.514$\pm$0.040 & 0.843$\pm$0.013 \\
HIVEN & 0.774$\pm$0.015 & 0.939$\pm$0.006 & \underline{0.760$\pm$0.008} & \underline{0.895$\pm$0.008} & 0.748$\pm$0.006 & 0.939$\pm$0.006 & 0.546$\pm$0.012 & 0.873$\pm$0.022 \\
SKES & 0.772$\pm$0.007 & 0.940$\pm$0.004 & 0.752$\pm$0.011 & 0.890$\pm$0.014 & 0.779$\pm$0.006 & 0.936$\pm$0.003 &0.539$\pm$0.012 &  0.885$\pm$0.004 \\
{LOGIC$^{\dagger}$} & {0.779$\pm$0.012}   &  {0.939$\pm$0.013}    & {0.751$\pm$0.024}   & {0.891$\pm$0.034}   & {0.737$\pm$0.011}   & {0.934$\pm$0.007}   & {0.540$\pm$0.011}   & {0.864$\pm$0.004 } \\ 
{SAT$^{\dagger}$}     & {0.761$\pm$0.007}  &  {0.924$\pm$0.009}   & {0.657$\pm$0.013 } & {0.873$\pm$0.004}  & {0.746$\pm$0.014}  & {0.917$\pm$0.034}  & {0.547$\pm$0.057}  & {0.883$\pm$0.017} \\ 
\hline
HGNN & 0.737$\pm$0.003 & 0.910$\pm$0.002 & 0.661$\pm$0.040 & 0.860$\pm$0.047 & 0.710$\pm$0.010 & 0.890$\pm$0.015 & 0.520$\pm$0.030 & 0.865$\pm$0.015 \\
HyperGAT & 0.769$\pm$0.010 & 0.932$\pm$0.013 & 0.675$\pm$0.010 & 0.869$\pm$0.030 & 0.760$\pm$0.010 & 0.920$\pm$0.010 & 0.550$\pm$0.028 & 0.875$\pm$0.010 \\
AllSetT & 0.765$\pm$0.011 & 0.938$\pm$0.013 & 0.720$\pm$0.013 & 0.887$\pm$0.002 & 0.770$\pm$0.012 & 0.928$\pm$0.012 & 0.543$\pm$0.030 & \underline{0.891$\pm$0.010} \\
ID-HAN & 0.772$\pm$0.009 & {0.940$\pm$0.010} & 0.746$\pm$0.020 & 0.893$\pm$0.004 & 0.765$\pm$0.017 & 0.939$\pm$0.014 & \underline{0.550$\pm$0.030} &  {0.880$\pm$0.012} \\
DualHGNN$^{\dagger}$ & 0.772$\pm$0.008 & 0.934$\pm$0.011 & 0.742$\pm$0.014 & 0.870$\pm$0.009 & 0.759$\pm$0.010 & 0.932$\pm$0.010 & 0.540$\pm$0.025 & 0.874$\pm$0.012 \\
DPHGNN$^{\dagger}$ & 0.746$\pm$0.012 & 0.923$\pm$0.014 & 0.747$\pm$0.005 & 0.891$\pm$0.010 & 0.769$\pm$0.009 & 0.934$\pm$0.007
& 0.534$\pm$0.008 & 0.871$\pm$0.013 
\\
DVHGNN$^{\dagger}$ & \underline{0.779$\pm$0.006} & 0.939$\pm$0.009& 0.758$\pm$0.013 & 0.882$\pm$0.006 & \underline{0.782$\pm$0.005} & \underline{0.940$\pm$0.003} & 0.545$\pm$0.014 &  {0.889$\pm$0.003} \\
\midrule
 \textbf{DualHNIE$^{\dagger}$} & \textbf{0.787$\pm$0.004} & \textbf{0.943$\pm$0.008} & \textbf{0.762$\pm$0.003} & \textbf{0.896$\pm$0.002} & \textbf{0.793$\pm$0.004} & \textbf{0.942$\pm$0.002} & \textbf{0.552$\pm$0.034} & \textbf{0.898$\pm$0.005} \\
\bottomrule \bottomrule 
\end{tabular}
\end{table*}

\subsection{Implementation Setup} 
\paragraph{Datasets}
We evaluate DualHNIE on four real-world heterogeneous graphs: FB15K, TMDB5K, MUSIC10K, and IMDB. The dataset statistics of heterogeneous knowledge graph and heterogeneous higher-order knowledge graph are summarized in Table~\ref{datasets}. we divide the training, validation, and test sets using a fixed 7:1:2 ratio.

\textbf{FB15K} is a subset of FreeBase\footnote[1]{http://www.freebase.be/}. It contains rich heterogeneous knowledge, including relational and textual information. Node importance labels are derived from the corresponding Wikipedia page views over the past 30 days.

\textbf{TMDB5K} is derived from the TMDB movie database\footnote[2]{https://www.kaggle.com/tmdb/tmdb-movie-metadata}, containing heterogeneous nodes such as actors, crew members, and companies. Semantic information comes from movie overviews, and node importance is annotated using the official movie popularity rating.

\textbf{IMDB} is processed from the IMDB database \footnote[3]{https://www.imdb.com/interfaces/}, whose nodes contain heterogeneous items such as movies, genres, casts, crews, publishing companies, and countries. The text information of this dataset comes from IMDB movie synopsis and personal biographies. The node importance labels are derived from IMDB movie votes.

\textbf{MUSIC10K} is a music database built on the 10k song datase
and supplemented with data from the Million Song Dataset\footnote[4]{https://millionsongdataset.com/}
. It contains about 22,985 entities, including artists, songs, artist terms. Since descriptive text is unavailable, entity names are used as semantic features. Artist familiarity defines the importance of artist nodes.

\paragraph{Baseline Methods}
To evaluate the effectiveness of DualHNIE, we compare it against a diverse set of state-of-the-art methods for node importance estimation, including traditional machine learning models, graph neural networks, and hypergraph neural networks. {Traditional baselines include PPR~\cite{haveliwala2002topic} and HAR~\cite{li2012har}. Among GNN-based methods, we consider GCN~\cite{kipf2016semi}, GraphSAGE~\cite{hamilton2017inductive}, GENI~\cite{park2019estimating}, LICAP~\cite{zhang2024label}, HIVEN~\cite{huang2022estimating}, and SKES~\cite{chen2024deep}, which leverage structural information in heterogeneous knowledge graphs to abstract multiple entities and relations. To ensure fairness, these structure-oriented methods are evaluated using only structural features.}
{Furthermore, we incorporate LLM-integrated frameworks, such as LOGIC~\cite{pan2025logic} and SAT~\cite{liu2025enhancing}, evaluated under identical semantic input conditions. These models allow us to assess DualHNIE’s structural-semantic decoupling performs relative to architectures that leverage the semantic understanding of Large Language Models.
To assess performance on higher-order relational modeling, we include hypergraph-based methods, such as structure-only methods HGNN~\cite{gao2022hgnn+}, HyperGAT~\cite{ding2020more}, and ID-HAN~\cite{chen2024hyperedge}. Moreover, we select dual-channels modelling methods with both structural features and semantic information, such as 
DualHGNN~\cite{wang2021dualgnn}, DPHGNN~\cite{saxena2024dphgnn}, and DVHGNN~\cite{li2025dvhgnn}. 
we evaluate baseline methods with two ranking-oriented metrics~\cite{park2019estimating} to capture different dimensions of ranking quality. NDCG evaluates the retrieval quality of top-ranked nodes, and Spearman correlation measures the consistency of the global ranking sequence.}

\paragraph{Implementation Details} 
\label{sectiona3}
All experiments are implemented in PyTorch and conducted on two NVIDIA RTX 6000 GPUs (48 GB). Models are trained for up to 10,000 epochs with early stopping (patience = 2000 epochs) and evaluated using 3-fold cross-validation for robustness. The default model consists of a single hidden layer with hidden dimension=20 and an attention mechanism with a dropout rate of 0.3. An adaptive fusion gate aligns structural and semantic representations, with contrastive regularization size fixed at 2000. {Multi-modal fusion parameters are initialized as $\eta_1$=0.3, $\eta_2$=0.7 and the regularization loss weights as $\alpha$=0.1, $\beta$=0.2, these weights are trainable parameters}. The model is optimized using the Adam optimizer with a learning rate of 5e-3 and weight decay of 5e-4.

\subsection{Experimental Results}

\subsubsection{Performance Evaluation on Node Importance Estimation}
Table~\ref{table 02} reports the performance of DualHNIE on heterogeneous higher-order knowledge graph compared with graph neural networks, hypergraph neural networks, and conventional machine learning methods. All results are averaged over 3-fold cross-validation, with mean values and standard deviations reported. The best results are highlighted in bold, and the second-best results are underlined.

{The experimental results validate that the proposed heterogeneous higher-order knowledge graph effectively bridges the gap inherent in traditional pairwise modeling, demonstrating substantial advantages in node importance estimation.} Specifically, on TMDB5K, DualHNIE consistently outperforms all GNN-based baselines (Spearman=0.762, NDCG@100=0.896), surpassing the runner-up model, ID-HAN (Spearman=0.746), by a margin of 2.21\%. On the IMDB dataset, DualHNIE maintains this competitive edge (Spearman=0.793, NDCG@100=0.942). Compared to specialized GNN models such as SKES (Spearman=0.779, NDCG@100=0.936) and LICAP (Spearman=0.760, NDCG@100=0.920), our framework achieves relative gains of 1.80\%/0.21\% and 4.34\%/2.39\%, respectively. Such performance underscores its proficiency in capturing complex semantic relationships through higher-order dependencies. {These findings confirm that DualHNIE delivers superior ranking performance across diverse benchmarks, yielding measurable improvements over both GNN and HGNN baselines. Its dual-branch architecture successfully leverages the higher-order knowledge graph to capture the intricate, multi-modal influence patterns that govern node importance in large-scale networks.}

\begin{table*}[h]
\small 
\centering
\setlength{\tabcolsep}{3pt} 
\caption{Evaluation of different channel in DualHNIE.}
\label{table03}
\begin{tabular}{lllllllll}
\toprule \toprule
\multirow{2}{*}{DualHNIE} 
& \multicolumn{2}{c}{FB15K} 
& \multicolumn{2}{c}{TMDB5K} 
& \multicolumn{2}{c}{IMDB} 
& \multicolumn{2}{c}{MUSIC10K} \\
\cmidrule(lr){2-3} \cmidrule(lr){4-5} \cmidrule(lr){6-7} \cmidrule(lr){8-9}
& SPEARMAN & NDCG@100 
& SPEARMAN & NDCG@100 
& SPEARMAN & NDCG@100 
& SPEARMAN & NDCG@100 \\
\toprule
DualHNIE-structure
& 0.720$\pm$0.002 & 0.914$\pm$0.003 
& \underline{0.759$\pm$0.003} & \underline{0.890$\pm$0.004} 
& 0.651$\pm$0.000 & \underline{0.934$\pm$0.008} 
& 0.395$\pm$0.054 & 0.826$\pm$0.005 \\
DualHNIE-semantic 
& 0.709$\pm$0.004 & 0.917$\pm$0.018 
& 0.613$\pm$0.018 & 0.778$\pm$0.021 
& \underline{0.703$\pm$0.002} & 0.940$\pm$0.004 
& \underline{0.512$\pm$0.007} & \underline{0.862$\pm$0.007} \\
DualHNIE-concat 
& \underline{0.731$\pm$0.003} & \underline{0.921$\pm$0.014} 
& 0.739$\pm$0.009 & 0.883$\pm$0.001 
& 0.726$\pm$0.002 & \underline{0.941$\pm$0.005} 
& 0.526$\pm$0.001 & 0.867$\pm$0.007 \\
\textbf{DualHNIE} 
& \textbf{0.787$\pm$0.004} & \textbf{0.943$\pm$0.003} 
& \textbf{0.762$\pm$0.003} & \textbf{0.896$\pm$0.002} 
& \textbf{0.793$\pm$0.004} & \textbf{0.942$\pm$0.002} 
& \textbf{0.552$\pm$0.034} & \textbf{0.898$\pm$0.005} \\
\bottomrule \bottomrule 
\end{tabular}
\end{table*}

\begin{table*}[htbp]
\small
\centering
\setlength{\tabcolsep}{3.5pt} 
\caption{Performance of DualHNIE variants comparison with different levels (NDCG@k).}
\label{table045}
\begin{tabular}{lllllllllllllllll}
\toprule \toprule
 & \multicolumn{4}{c}{FB15K NDCG@k} & \multicolumn{4}{c}{TMDB5K NDCG@k} & \multicolumn{4}{c}{IMDB NDCG@k} & \multicolumn{4}{c}{MUSIC10K NDCG@k} \\
 \cmidrule(lr){2-17}
 & 20 & 50 & 100 & 200 & 20 & 50 & 100 & 200 & 20 & 50 & 100 & 200 & 20 & 50 & 100 & 200 \\
 \toprule
DualHNIE-structure & 0.882 & 0.903 & 0.915 & \underline{0.925} & \underline{0.866} & \underline{0.882} &  \underline{0.890} & \underline{0.896} & 0.908 & \underline{0.920} & 0.929 & \underline{0.940} & 0.869 & 0.826 & 0.826 & 0.853 \\
DualHNIE-semantic & 0.865 & 0.898 & \underline{0.917} & 0.927 & 0.703 & 0.790 & 0.793 & 0.812 & \underline{0.901} & \underline{0.916} & \underline{0.925} & 0.937 & \underline{0.892} & \textbf{0.874} & \textbf{0.862} & \textbf{0.878} \\
DualHNIE-concat & \underline{0.891} & \underline{0.906} & 0.921 & \underline{0.938} & 0.865 & 0.873 & \underline{0.883} & \textbf{0.901} & \underline{0.909} & 0.920 & \underline{0.930} & \underline{0.940} & 0.859 & 0.826 & 0.821 & \underline{0.855} \\
DualHNIE & \textbf{0.905} & \textbf{0.921} & \textbf{0.943} & \textbf{0.941} & \textbf{0.873} & \textbf{0.876} & \textbf{0.896} & \underline{0.896} & \textbf{0.912} & \textbf{0.923} & \textbf{0.933} & \textbf{0.942} & \textbf{0.896} & \underline{0.867} & \underline{0.858} & \textbf{0.878} \\
\bottomrule \bottomrule 
\end{tabular}
\end{table*}

\subsubsection{Performance evaluation on both structural and semantic channels} Table \ref{table03} and \ref{table045} display the proposed dual-channel design in DualHNIE contributes to overall model performance, {we conduct an ablation study by comparing four architectural variants at different NDCG$@k$ with $k $=$20,50,100,200$.} The variants include: (1) DualHNIE-\textit{structure}, a structural prediction score with hypergraph attention branch; (2) DualHNIE-\textit{semantic}, a semantic prediction score using sparse-chunked hypergraph transformer branch; (3) DualHNIE-\textit{concat}, which combines structural and semantic feature concatenation; and (4) \textit{DualHNIE}, which integrates dual-branch processing.

Table~\ref{table03} reports DualHNIE consistently achieves superior performance, validating the effectiveness of its dual-branch architecture that explicitly disentangles and integrates both structural and semantic information. {On FB15K, DualHNIE achieves a Spearman score of 0.787 and NDCG@100 of 0.943, representing improvements of 9.31\% and 3.17\% over the structure-only model, and 9.74\% and 2.84\% over the semantic-only model, respectively.} Compared to the concatenation-based variant, it yields gains of 6.43\% in Spearman and 2.39\% in NDCG@100, confirming that structured coordination outperforms simple feature fusion.
For TMDB5K, the semantic-only variant performs substantially worse than the structure-only model (23.84\% lower in Spearman and 14.40\% in NDCG@100). Nonetheless, DualHNIE achieves the best performance (Spearman:0.737, NDCG@100:0.896), improving NDCG@100 by 3.10\% over the best single-branch variant.
These results consistently support that explicit disentanglement and joint modeling of structure and semantics significantly enhance node importance estimation.

Table \ref{table045} shows that DualHNIE demonstrates particularly strong performance at small values of $k$ (e.g., 20 and 50), outperforming all baseline models. On FB15K, it achieves improvements over the best structural and semantic baselines, highlighting its enhanced accuracy in identifying the most critical top-ranked nodes. As $k$ increases to 100 and 200, DualHNIE maintains top-tier performance, confirming its robustness across varying recommendation depths. The model consistently ranks at or near state-of-the-art levels even with longer candidate lists, illustrating its stable integration of structural and semantic signals without degradation in ranking quality. These results demonstrate that DualHNIE delivers superior performance across all evaluation depths, affirming that its dual-channel architecture effectively enhances both top-k ranking accuracy.

\subsubsection{Heterogeneous High-order Knowledge Graph modeling}
{Table \ref{table02} evaluates the benefits of our dual-channel design for capturing high-order dependencies and decoupling structural semantic information,} we compare DualHNIE against dual-channel baselines (DualHGNN~\cite{wang2021dualgnn}, DVHGNN~\cite{li2025dvhgnn}, DPHGNN~\cite{saxena2024dphgnn}). We further include comparisons with dual-channel graph models (DualGAT, DualGT, DualGAT-GT) on HKGs and dual-channel hypergraph models (DualHGAT, DualHGT) on HHKGs .
{As illustrated in Table \ref{table02}, DualHNIE consistently outperforms DPHGNN and DVHGNN, achieving the highest scores on both Spearman correlation (0.787) and NDCG$@$100 (0.943). This indicates that our framework enables explicit and complementary modeling of semantically rich higher-order interactions.}
Compared to pairwise modeling of GNN methods, such as DualGAT, DualGT, DualGAT-GT on HKGs, DualHNIE shows notable improvements, which benefits from incorporating higher-order structures in heterogeneous graph modeling.
{Compared to the DualHGAT and DualHGT models, DualHNIE with two-view modules achieves superior performance, highlighting the benefits of structural and semantic decoupling modeling and alignment fusion. These results indicate that heterogeneous higher-order knowledge graphs and disentangled learning more effectively capture complex structural and semantic relations between multi-users and items. }

\begin{table}[ht]
\centering
\small
\setlength{\tabcolsep}{1pt} 
\caption{Performance Comparison of dual-channel models. }
\label{table02}
\begin{tabular}{lcccc}
\toprule \toprule
 & \multicolumn{2}{c}{FB15K} & \multicolumn{2}{c}{MUSIC10K} \\
 \cmidrule(lr){2-5}
 & 
 SPEARMAN &
 NDCG@100 &
 SPEARMAN &
 NDCG@100 \\
 \midrule
DualHGNN & 0.772$\pm$0.008 & 0.934$\pm$0.011 & 0.540$\pm$0.025 & 0.874$\pm$0.012 \\
DPHGNN & 0.746$\pm$0.012
 & 0.923$\pm$0.014 
 & 0.534$\pm$0.008 
 & 0.871$\pm$0.013 \\
DVHGNN & 0.779$\pm$0.006 
& 0.939$\pm$0.009 
& \underline{0.545$\pm$0.014}
& \underline{0.889$\pm$0.003}
\\
\midrule
DualGAT &
 0.772$\pm$0.018 &
 0.942$\pm$0.005 &
 0.439$\pm$0.010 &
 0.843$\pm$0.028 \\
DualGT &
 \underline{0.782$\pm$0.001} &
 0.934$\pm$0.013 &
 0.474$\pm$0.021 &
 0.865$\pm$0.014 \\
DualGAT-GT &
 0.779$\pm$0.006 &
 \underline{0.938$\pm$0.009} &
 0.476$\pm$0.011 &
 0.863$\pm$0.018 \\
DualHGAT &
 0.703$\pm$0.012 &
 0.924$\pm$0.008 &
 0.417$\pm$0.014 &
 0.854$\pm$0.010 \\
DualHGT &
 0.765$\pm$0.013 &
 0.930$\pm$0.011 &
 0.471$\pm$0.021 & 
 0.711$\pm$0.019 \\
\textbf{DualHNIE} & \textbf{0.787$\pm$0.004} & \textbf{0.943$\pm$0.008} & \textbf{0.552$\pm$0.034} & \textbf{0.898$\pm$0.005} \\
\bottomrule \bottomrule
\end{tabular}
\end{table}

\subsection{Ablation Studies.}

\subsubsection{Ablation Study of Loss Regularization}
Table~\ref{tab:ablation-loss} presents an ablation study of the loss components using NDCG@100. The full mode (\(L_1+L_2\)) consistently achieves the highest performance across all datasets, demonstrating the complementary effects of the contrastive loss(\(L_1\)) and the unimodal prediction loss (\(L_2\)). 
Removing the contrastive loss (w/o-\(L_1\)) results in a substantial performance drop, e.g., from 0.898 to 0.852 on MUSIC10K, highlighting the critical role of contrastive learning in aligning structural and semantic representations. This effect is particularly pronounced on datasets with higher modality heterogeneity, where effective cross-modal alignment is essential. 
Meanwhile, excluding the unimodal loss (w/o-\(L_2\)) also leads to clear performance degradation, notably on FB15K and IMDB, indicating that unimodal supervision is necessary to preserve predictive capability within individual modalities and to prevent over-reliance on cross-modal signals. 
Interestingly, the variant without both loss components (w/o-\(L_1+L_2\))occasionally outperforms the single-loss variants, suggesting some partial redundancy in how each loss regularizes the model. Nevertheless, it still underperforms the full model, confirming that \(L_1\) and \(L_2\) jointly provide complementary regularization, ensuring effective cross-modal alignment while maintaining unimodal predictive fidelity. 

\begin{table}[htbp]
\caption{Ablation study of loss function (NDCG@100)}
\small
\label{tab:ablation-loss}
\setlength{\tabcolsep}{2pt}
\setlength{\abovecaptionskip}{0pt}
\setlength{\belowcaptionskip}{0pt} 
\begin{tabular}{lcccc}
\toprule \toprule\multirow{1}{*}{DualHNIE} 
 & FB15K & TMDB5K & IMDB & MUSIC10K \\
\midrule
w/o-$L_1$ & 0.927$\pm$0.004 & \underline{0.893$\pm$0.005} & 0.925$\pm$0.009 & 0.852$\pm$0.020 \\
w/o-$L_2$ & \underline{0.934$\pm$0.001} & 0.891$\pm$0.010 & 0.928$\pm$0.011 & \underline{0.868$\pm$0.003} \\
w/o-$L_1$+$L_2$ & 0.935$\pm$0.005 & 0.891$\pm$0.005 & \underline{0.934$\pm$0.004} & 0.861$\pm$0.012 \\
$L_1$+$L_2$ & \textbf{0.943$\pm$0.008} & \textbf{0.896$\pm$0.002} & \textbf{0.942$\pm$0.003} & \textbf{0.898$\pm$0.005} \\
\bottomrule \bottomrule 
\end{tabular}
\end{table}

\begin{table}[htbp]
\small
\caption{Evaluation of Sparse-Chunked Attention.}
\label{tab:performance}
\setlength{\tabcolsep}{2pt}
\setlength{\abovecaptionskip}{0pt}
\setlength{\belowcaptionskip}{0pt} 
\begin{tabular}{llcccc}
\toprule \toprule 
 & & SPEARMAN & NDCG@100 & Time/s & GPU/G \\
\hline
\multirow{2}{*}{FB15K} 
 & \textit{w/o} & 0.772$\pm$0.002 & 0.936$\pm$0.004 & 1.561 & 5.3 \\
 & ours & 0.778$\pm$0.005 & 0.943$\pm$0.003 & 0.239 & 2.7 {\fontsize{6pt}{6pt}\selectfont(+49.06\%)} \\
\hline
\multirow{2}{*}{TMDB5K} 
 & \textit{w/o} & 0.751$\pm$0.003 & 0.873$\pm$0.013 & 2.804 & 7.5 \\
 & ours & 0.762$\pm$0.003 & 0.896$\pm$0.002 & 0.518 & 3.3 {\fontsize{6pt}{6pt}\selectfont(+56.12\%)} \\
\hline
\multirow{2}{*}{IMDB} 
 & \textit{w/o} & 0.774$\pm$0.013 & 0.938$\pm$0.020 & 5.728 & 63.7 \\
 & ours & 0.793$\pm$0.004 & 0.942$\pm$0.002 & 3.561 & 34.8 {\fontsize{6pt}{6pt}\selectfont(+45.46\%)} \\
\hline
\multirow{2}{*}{MUSIC10K} 
 & \textit{w/o} & 0.497$\pm$0.015 & 0.845$\pm$0.021 & 0.702 & 1.8 \\
 & ours & 0.550$\pm$0.034 & 0.878$\pm$0.005 & 0.334 & 1.3 {\fontsize{6pt}{6pt}\selectfont(+28.41\%)} \\
\bottomrule \bottomrule
\end{tabular}
\end{table}

\subsubsection{Ablation Study of Sparse-Chunked Aggregation} 
To address the quadratic complexity of full attention between queries and keys in the hypergraph transformer, we introduce a Sparse-Chunked Aggregation strategy that enhances scalability on large datasets. As shown in Table~\ref{tab:performance}, incorporating this mechanism leads to consistent improvements across all four benchmarks in both SPEARMAN and NDCG@100 metrics. For instance, the most notable improvement occurs on MUSIC10K, where SPEARMAN rises from 0.497 to 0.550 ($+10.7\%$) and NDCG@100 from 0.845 to 0.878 ($+3.9\%$).
These results validate that the Sparse-Chunked Aggregation mechanism not only reduces computational overhead but also strengthens the model’s representational capacity across diverse datasets.
Moreover, Sparse-Chunked Aggregation also yields substantial efficiency gains. Results in Table~\ref{tab:performance} indicate a reduction in runtime of over 30\% across all datasets, with an 84.7\% decrease observed on FB15K (from 1.561s to 0.239s). 
GPU memory usage also decreases markedly. For instance, on FB15K, the runtime drops from 1.561 s to 0.239 s, and GPU memory consumption decreases by 49.06\%; on TMDB5K, memory usage falls by 56.12\%, enabling configurations that were previously infeasible due to memory constraints. These gains arise directly from chunked sparsification, which eliminates redundant computations and substantially reduces the storage overhead of large attention matrices, making large-scale hypergraph modeling tractable.


\subsubsection{Ablation Study on Gate Mechanism} 
We evaluate the effectiveness of different fusion strategies for integrating structural and semantic embeddings (Equation.~\ref{fusion}). Table~\ref{table-A6} displays \emph{adaptive gate} consistently achieves the best performance on FB15K and TMDB5K, with NDCG@100 of 0.94 and 0.886, and Spearman correlations of 0.787 and 0.760, respectively, outperforming attention-based, concatenation, fixed-weight summation, and simple gating mechanisms.
The adaptive gate’s superiority stems from its ability to dynamically balance structural and semantic contributions on a per-instance basis. Unlike static methods such as fixed-weight summation or concatenation, which may introduce redundancy or ignore input-specific variations, the adaptive gate emphasizes the more informative modality for each node. This flexibility is crucial in heterogeneous knowledge graphs, where the relative importance of structural and semantic signals varies across nodes and relations. Consistent gains across datasets and metrics confirm that adaptive gating provides the most effective fusion for DualHNIE.

\begin{table}[htbp]
\caption{Evaluation of different fusion mechanisms}
\small
\label{table-A6}
\setlength{\tabcolsep}{3pt} \setlength{\abovecaptionskip}{0pt}
\setlength{\belowcaptionskip}{0pt} 
\begin{tabular}{lllll}
\toprule \toprule
\multirow{2}{*}{} & \multicolumn{2}{c}{FB15K} & \multicolumn{2}{c}{TMDBK} \\
\cmidrule(lr){2-5}
 & NDCG@100 & SPEARMAN & NDCG@100 & SPEARMAN \\ 
\midrule
Adaptive & \textbf{0.943$\pm$0.005} & \textbf{0.787$\pm$0.003} & \textbf{0.886$\pm$0.003} & \textbf{0.760$\pm$0.004} \\
Attention & 0.920$\pm$0.003 & 0.749$\pm$0.016 & 0.860$\pm$0.017 & 0.733$\pm$0.013 \\
Concat & \underline{0.930$\pm$0.003} & \underline{0.771$\pm$0.012} & \underline{0.885$\pm$0.006} & \underline{0.747$\pm$0.009} \\
Fixed & 0.931$\pm$0.003 & 0.774$\pm$0.004 & 0.868$\pm$0.014 & 0.723$\pm$0.005 \\
Gate & 0.919$\pm$0.003 & 0.752$\pm$0.019 & 0.883$\pm$0.009 & 0.750$\pm$0.015\\
\bottomrule \bottomrule
\end{tabular}
\end{table}

\subsection{Parameter Analysis}

\noindent 
\subsubsection{Parameter Analysis of Fusion and Loss Function.} 
\begin{figure}[htbp]
\centering
\includegraphics[width=1\linewidth]{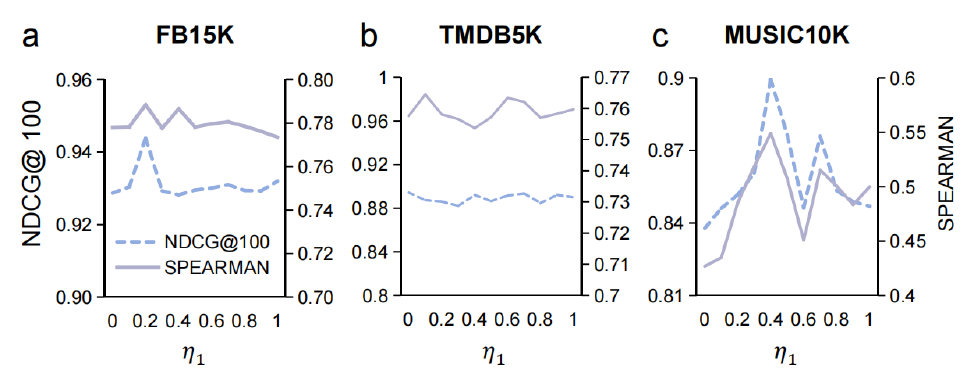}
\caption{Hyperparameter $\eta_1$ analysis of fusion ($\eta_2=1-\eta_1$). }
\label{figure06}
\end{figure}

\begin{figure}[htbp]
 \centering
 \includegraphics[width=1\linewidth]{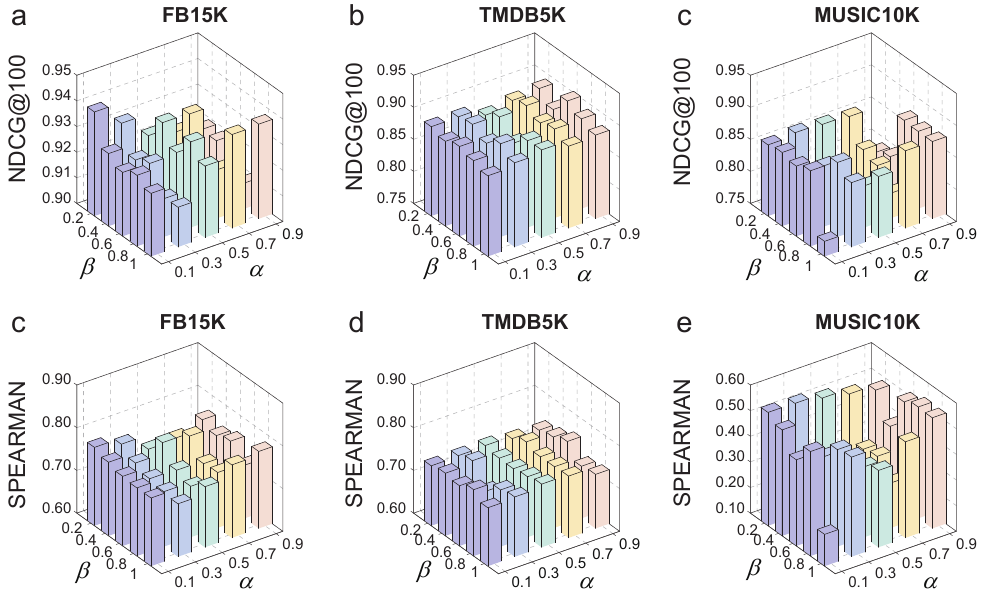}
 \caption{Hyperparameter analysis of loss function.}
 \label{figure07}
\end{figure}

{Fig. \ref{figure06}
presents an ablation study on the training fusion weight $\eta_1$ in Equation~\eqref{fusion} (with $\eta_2 = 1 - \eta_1$) to analyze the sensitivity of performance to the structural--semantic trade-off. 
on FB15K and TMDB5K, both NDCG$@$100 and Spearman correlation remain remarkably stable across $\eta_1 \in (0,1)$. However, when $\eta_1=0$ (semantic-only) and $\eta_1=1$ (structure-only), the results of the model deteriorated, indicating that structural and semantic signals are complementary and aligned for node importance estimation.  
In contrast, MUSIC10K exhibits a distinct performance peak at $\eta_1 = 0.4(\eta_2=0.6)$, implying that the model relies more heavily on semantic cues (e.g., music descriptions). These results indicate that our flexibly trained fusion gate enables the model to balance the inherent signals learned from each knowledge graph.}
We further analyze the loss weighting parameters $\alpha$ and $\beta$ (Equation.~\ref{fusionloss}), as shown in Fig. \ref{figure07}. Results indicate that a small to moderate $\alpha$ (0.1--0.5) generally yields higher NDCG@100 and Spearman values, implying that light contrastive regularization is sufficient to align cross-modal representations without impairing unimodal performance. Similarly, intermediate values of $\beta$ (0.2--0.6) lead to robust outcomes. For instance, the combination of small $\alpha$ and moderate $\beta$ achieves the best trade-off across datasets, e.g., $(\alpha=0.1, \beta=0.2)$ on FB15K, $(\alpha=0.3, \beta=0.4)$ on TMDB5K, and $(\alpha=0.1, \beta=0.2)$ on MUSIC10K.

\subsubsection{Parameter Analysis of Layers Number and Attention Heads.}
\begin{figure}[htbp]
 \centering
 \includegraphics[width=1\linewidth]{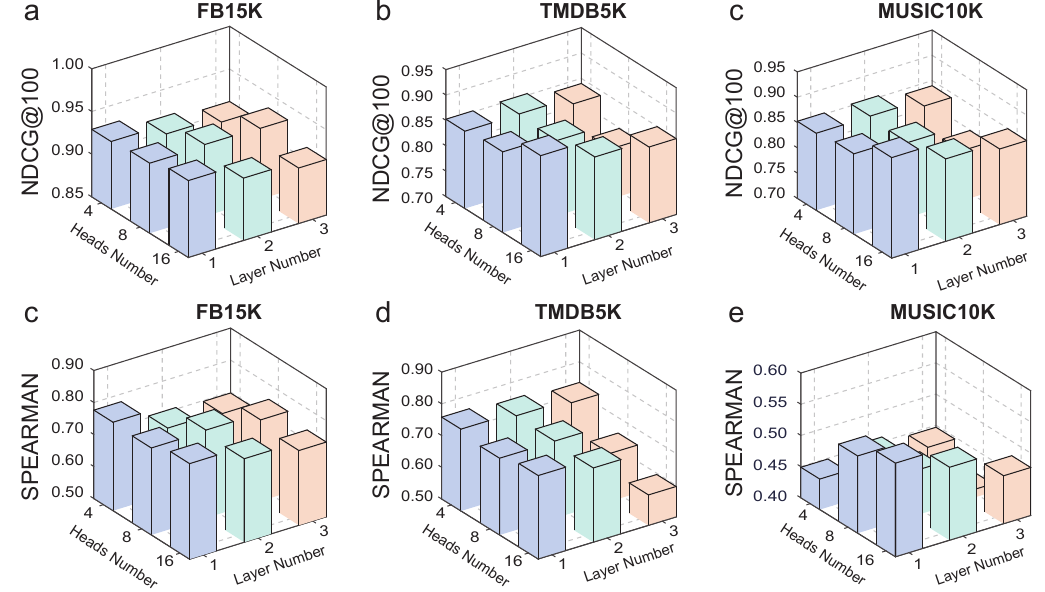}
 \caption{Ablation study of layer depth and heads' number.}
 \label{figure05}
\end{figure}

We analyze the effect of the number of model layers and attention heads on performance in Fig.~\ref{figure05}.
Increasing the number of layers leads to performance degradation. On FB15K, both NDCG@100 and Spearman remain stable with 1--2 layers but degrade notably with 3 layers (e.g., NDCG@100 drops from 0.90 to 0.79 with 16 heads; Spearman decreases from 0.80 to 0.60). On TMDB5K: 2 layers can provide slight gains over a single layer, but models with 3 layers consistently underperform. This suggests that deeper layer architectures may suffer from over-smoothing.
The number of attention heads generally improve performance but at increased computational cost. On FB15K, 4--16 heads generally yield stable performance. Similarly, On TMDB5K the model achieves the best performance with 1 layer and 16 heads. However, when model combining deeper layers with more heads diminishes these gains.

\subsection{Complexity Comparison} 
\begin{figure}[htbp]
 \centering
\includegraphics[width=1\linewidth]{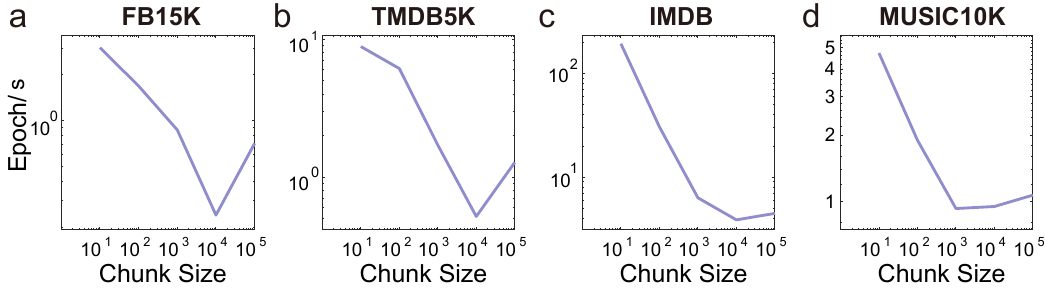}
 \caption{Runtime(/s) of SCA with chunk size(10$^1$ to 10$^6$).}
 \label{figureA1}
\end{figure}

Table~\ref{tableA-3} compares the GPU memory consumption of the structural and semantic channels across four models: the graph attention network GNIE~\cite{park2019estimating}, the graph Transformer RGTN~\cite{huang2021representation}, DualHNIE without SCA, and DualHNIE with SCA, under identical training settings (Table~\ref{tableA-3}). On FB15K, DualHNIE(SCA) reduces GPU usage to 0.6 (structural) and 1.1 (semantic), saving 65--80\% compared to GNIE (1.8, 2.8) and RGTN (3.5, 4.7). On IMDB, savings remain substantial: SCA lowers structural/semantic costs from 41.3/46.5 (GNIE) and 53.1/55.9 seconds (RGTN) to 26.1/32.3, achieving 35--50\% reduction.
Figure~\ref{figureA1} shows that runtime decreases as the chunk size $C$ increases, reaching a minimum at a moderate value ($C=10{,}000$), but increases again for large chunks ($C=100{,}000$). Small chunks incur high overhead due to frequent scattering operations, while moderate chunk sizes enable efficient computation. 
These results demonstrate that SCA effectively reduces computational cost by processing only nonzero entries, allowing DualHNIE to maintain representational capacity while achieving notable efficiency gains.

\begin{table}[htbp]
\caption{GPU memory consumption (GPU/GB). Best values (\textbf{Bold}) and second-best (\underline{Underlined}).}
\label{tableA-3}
\small
\setlength{\tabcolsep}{1pt} 
\begin{tabular}{lcccccc}
\toprule \toprule
 & \multicolumn{2}{c}{FB15K} & \multicolumn{2}{c}{TMDB5K} & \multicolumn{2}{c}{IMDB} \\
 \cmidrule(lr){2-7}
 & 
 \multicolumn{1}{c}{struct} &
 \multicolumn{1}{c}{semantic} &
 \multicolumn{1}{c}{struct} &
 \multicolumn{1}{c}{semantic} &
 \multicolumn{1}{c}{struct} &
 \multicolumn{1}{c}{semantic} \\
 \hline
GNIE & \underline{1.8} & \underline{2.8} & 3.0 & 10.7 & 41.3 & 46.5 \\
RGTN & 3.5 & 4.7 & 4.7 & 11.7 & 53.1 & 55.9 \\
DualHNIE(w/o-SCA) & 2.6 & 3.8 & \underline{2.3} & \underline{5.6} & \underline{39.8} & \underline{48.2} \\
DualHNIE(SCA) & \textbf{0.6} & \textbf{1.1} & \textbf{2.0} & \textbf{3.1} & \textbf{26.1} & \textbf{32.3} \\
\bottomrule\bottomrule
\end{tabular}
\end{table}

 \section{Conclusion}

In this paper, we propose DualHNIE, a dual-channel hypergraph learning framework for node importance estimation in heterogeneous knowledge graphs. By constructing meta-path--induced hypergraphs, DualHNIE models collective, high-order relational patterns that are inaccessible to conventional pairwise message-passing schemes. Its dual-channel architecture explicitly disentangles structural topology from semantic context, yielding complementary and more interpretable representations, while the proposed sparse--chunked hypergraph transformer enables efficient computation on large-scale knowledge graphs.
Extensive experiments on multiple NIE benchmarks demonstrate that DualHNIE consistently improves over state-of-the-art approaches. These results demonstrate that DualHNIE achieves superior performance on node importance estimation, validating the effectiveness of modeling higher-order dependencies and integrating structure--semantics information in heterogeneous knowledge graphs.
In the future, DualHNIE can be extended to downstream tasks, such as personalized recommendation, intelligent search and ranking in complex networked applications.

\bibliographystyle{IEEEtran}
\bibliography{IEEEtran}

\end{document}